\documentclass[10pt,twocolumn,letterpaper]{article}

\usepackage{cvpr}              

\definecolor{cvprblue}{rgb}{0.21,0.49,0.74}
\usepackage[pagebackref,breaklinks,colorlinks,allcolors=cvprblue]{hyperref}
\usepackage{multirow,colortbl}
\usepackage{pifont}
\usepackage{lipsum}
\def\confName{CVPR}

\definecolor{gray0}{gray}{0.9}

\title{A dual contrastive framework}

\author{
    Yuan Sun \\
    Rutgers University \\
    {\tt\small ys820@rutgers.edu}
    \and
    Zhao Zhang \\
    Rutgers University \\
    {\tt\small zz671@soe.rutgers.edu}
    \and
    Jorge Ortiz \\
    Rutgers University \\
    {\tt\small jorge.ortiz@rutgers.edu}
}

\begin{document}
\maketitle
\begin{abstract}

 \sloppy In current multimodal tasks, models typically freeze the encoder and decoder while adapting intermediate layers to task-specific goals, such as region captioning. Region-level visual understanding presents significant challenges for large-scale vision-language models. While limited spatial awareness is a known issue, coarse-grained pretraining, in particular, exacerbates the difficulty of optimizing latent representations for effective encoder-decoder alignment. We propose AlignCap, a framework designed to enhance region-level understanding through fine-grained alignment of latent spaces. Our approach introduces a novel latent feature refinement module that enhances conditioned latent space representations to improve region-level captioning performance. We also propose an innovative alignment strategy, the semantic space alignment module, which boosts the quality of multimodal representations.
Additionally, we incorporate contrastive learning in a novel manner within both modules to further enhance region-level captioning performance. To address spatial limitations, we employ a General Object Detection (GOD) method as a data preprocessing pipeline that enhances spatial reasoning at the regional level. Extensive experiments demonstrate that our approach significantly improves region-level captioning performance across various tasks.
\end{abstract}    
\section{Introduction}
\label{sec:intro}

\begin{figure}[t]
  \centering
   \includegraphics[width=1\linewidth]{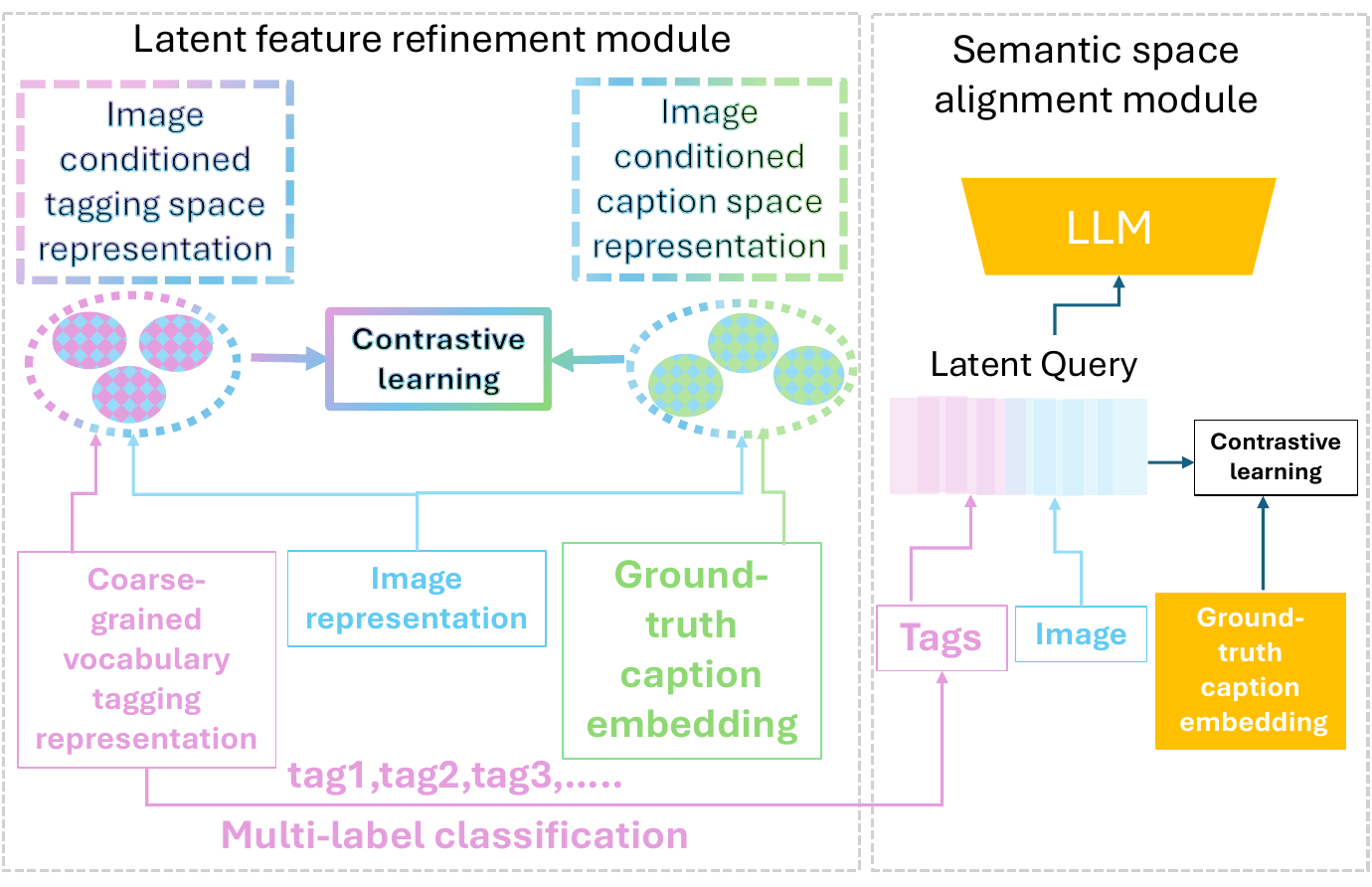}
   \caption{Summary of the refinement module in AlignCap, which refines latent features through a dual contrastive pipeline.}
   \label{fig:contra_module}
   \vspace{-4mm}
\end{figure}

LLMs have exhibited remarkable abilities in leveraging prior knowledge and reasoning, stimulating significant interest in expanding these capabilities to other modalities. This has driven the development of Multimodal Large Language Models (MLLMs), which are able to process and interpret information across multiple input sources beyond text. With the progression of MLLMs, they have been applied to numerous downstream tasks. These models demonstrate strong performance on image-level tasks like classification, image captioning, coarse-grained retrieval, and visual question answering. However, these models often discard fine-grained latent embeddings, resulting in suboptimal performance on tasks requiring the understanding of spatial relationships between objects or object attributes. These limitations are particularly pronounced when such pre-trained models are employed as foundation models.

Region-level captioning demands fine-grained descriptions and a deeper understanding of specific image areas. While existing methods often transform image regions into pure text, they tend to overlook the rich vision-domain information necessary for image-to-text descriptions. Recent works address this by leveraging visual representations for region-level captioning, whereas others focus on improving performance through novel architectural enhancements in language models. To balance computational efficiency with accuracy, many current approaches adopt task-specific models by bootstrapping from pre-trained vision and language models.

However, most current methods focus on alignment at a coarse-grained level, where communication between modalities relies directly on pre-trained latent representations from distinct signals. For instance, many popular LVLMs utilize Vision Transformers (ViT) pre-trained via CLIP as image encoders. While these visual features excel at pre-defined label classification and image-text matching, they are confined to a pre-defined space, often lacking task-specific adaptability. This presents challenges when applying such pre-trained features to domain-specific tasks. Moreover, since LLMs are trained in unimodal settings and have not encountered images, freezing them during multimodal training complicates vision-language alignment. Finally, existing methods typically operate over fixed image regions, limiting the model's ability to adapt to intricate region details and capture rich contextual information.

We propose \textit{AlignCap}, a method that addresses fine-grained multimodal alignment (Fig.~\ref{fig:contra_module}) for region-level captioning, enhancing both the captioning quality and spatial understanding. Most current approaches rely on coarse-grained closed-set classification tasks, where human-annotated class names are fed into prompts to generate captions. However, this comprehensive vocabulary tagging space, which encompasses all possible classes, is not suitable for traditional contrastive learning, as it requires specific text tailored to the target image. Our \textbf{\textit{latent feature refinement} }module addresses this innovatively by aligning image-conditioned latent representations across different spaces. Specifically, the coarse-grained vocabulary tagging space is conditioned on the visual signal to be refined into a fine-grained space. Similarly, we conditioned the image on the ground-truth caption to produce image-conditioned caption space representations. Contrastive learning between these two image-conditioned representations refines both image and text features, transforming a coarse-grained representation into a fine-grained one, thereby improving region-level captioning. Additionally, to resolve alignment issues introduced by freezing the LLM during multimodal training, we propose a novel strategy, the \textbf{\textit{semantic space alignment}} module, to align multimodal inputs with the pre-trained latent space of a frozen LLM, ensuring the LLM can effectively utilize intermediate representations. Finally, a distinctive \textbf{\textit{General Object Detection }}(GOD) pipeline enhances spatial awareness by proposing multiple objects of interest, which are processed alongside the target region to improve the model's ability to understand spatially-aware semantic concepts for enhanced object recognition and captioning.

To summarize, the key contributions of this work are as follows:

\begin{itemize}

    \item We introduce \textit{AlignCap}, a novel approach for optimizing the latent representations of MLLMs, significantly enhancing both region-level captioning quality and spatial comprehension. Our method achieves substantial performance improvements over state-of-the-art baselines in experimental evaluations.
     
    \item We develop a dual contrastive learning framework (Fig.~\ref{fig:contra_module}) featuring two novel modules to refine the latent representations of MLLMs, enhancing region-level captioning. First, we introduce a \textit{Latent Feature Refinement} module that transforms coarse-grained latent representations into fine-grained, image-conditioned representations. Second, we propose a novel \textit{Semantic Space Alignment} module, which optimizes the input representations for the frozen LLM decoder, improving its capacity to process and generate captions.
    
    \item We implement a distinctive pipeline incorporating the \textit{General Object Detection} (GOD) method to detect general objects of interest, boosting the model's spatial understanding and object recognition capabilities for improved region-level captioning.

\end{itemize}
\section{Related Work}
\label{sec:ReWo}

\subsection{Region-level Image Comprehension}

Region-level computer vision tasks have long been central to research. Traditional object detection identifies candidate regions and classifies their contents, while open-vocabulary recognition expands categories using natural language. Semantic segmentation achieves pixel-wise classification within a closed-set paradigm. Region-level captioning, however, provides more context-sensitive insights. Modern models leverage LLMs to generate region-specific interpretations, such as RegionGPT, which uses spatial-aware modules for tasks like captioning. VisionLLM and GPT4RoI align region features with LLM embeddings, allowing fine-grained descriptions. Other approaches enhance specificity using human feedback or reference highlighting. For instance, ASM uses human feedback and semantic tags to improve multi-task performance. ControlCap addresses caption degeneration with control words from human feedback, enabling detailed captions. Yet, aligning multimodal inputs for enhanced region captioning remains a key challenge.

\subsection{Latent Representation Alignment}

Traditional methods align discriminative representations across modalities using contrastive learning. These methods employ a visual encoder and a text encoder to independently project visual and text inputs into a shared representation space. While these approaches focus on modality-specific signals, recent research has shifted towards examining latent variables conditioned on information across modalities. Joint-embedding predictive architectures (JEPA) infer the representation from one modality based on another, utilizing a context variable to describe the transformation from the source to the target modality. Similarly, latent Language Image Pretraining (Llip) generates a visual representation conditioned on a text caption through contrastive learning.

Recent works explore dual and multi-contrastive frameworks to enhance multimodal alignment. Building on these, our method refines coarse-grained features via contrastive learning, generating fine-grained features aligned with specific semantics. Our coarse-grained representation lacks a direct pairing and is conditioned on images. We employ a dual-contrastive framework to enhance region-level captioning performance.

\subsection{General Object Detection}

Region proposal methods are crucial in tasks like object detection and segmentation. While MLLMs excel in image captioning, they struggle with accurate object localization. We propose a selective region proposal approach (GOD) within the MLLM framework to enhance region-level captioning by improving spatial awareness.

\section{Methodology}
\label{sec:Meth}

\begin{figure*}[tbph!]
  \centering
   \includegraphics[width=1\linewidth]{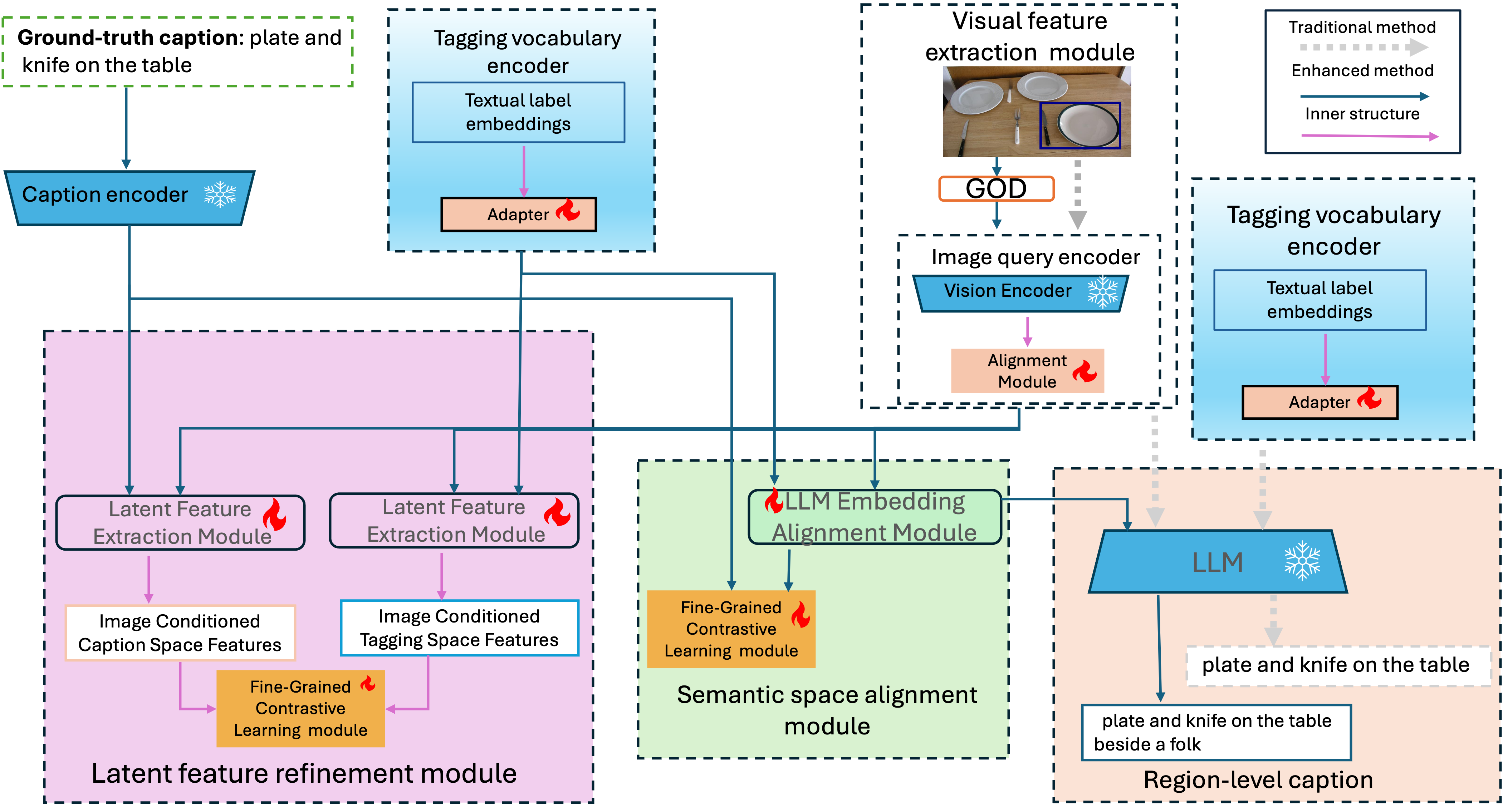}
\caption{Overview of the proposed AlignCap architecture. Following a conventional design that extracts semantic visual evidence as keyword tags and sends the latent image query together to get the caption from an LLM, we design the Semantic Space Alignment Module and Latent Feature Refinement Module to enhance the performance of multimodal representation. We also introduce a GOD Module in the visual feature extraction stage to enhance the spatial awareness of our region-level captioning model.}
   \label{fig:overview_arch}
   \vspace{-3.5mm}
\end{figure*}

\subsection{Model Structure Overview}

\noindent AlignCap (Fig.~\ref{fig:overview_arch}) follows the mainstream paradigm of bootstrapping vision-language pre-training from off-the-shelf, frozen pre-trained image encoders and large language models (LLMs). The model adopts a conventional design, consisting of a trainable vision sampler to extract image features from a frozen ViT encoder, and a lightweight tagging space encoder that bootstraps latent representations from a frozen CLIP encoder. This tagging space encoder provides tags from subclasses such as entities, attributes, actions, and scenes via discriminative learning(detailed in  Appendix Section). Finally, a frozen LLM generates region-level captions from the latent representations of the tags and the image.

\noindent As discussed in the previous section, conventional methods encounter several limitations. To address these issues, we initially introduce a \textbf{\textit{General Object Detection}}(GOD) module. By jointly learning the regions of the proposed objects from GOD alongside the target region through the \textbf{\textit{visual feature extraction}} module, our approach enriches the spatial context for the target region, thereby improving spatial understanding. After extracting visual features from the input image, instead of directly passing them to the LLM with identified words, we propose two novel modules to refine the image and text features for region-level captioning.

First, we refine the latent features via contrastive learning in the \textbf{\textit{latent feature refinement}} module. While traditional methods employ image-text pair for contrastive learning to enhance discriminative learning, our method leverages a tagging space embedding, which encodes a broad vocabulary, possibly including which might be of varying relevance to the current image. The primary function of the vocabulary tagging embedding is to identify the most relevant tags from this expansive vocabulary based on the image embeddings. This coarse-grained vocabulary tagging embedding, created by a frozen encoder pre-trained on a general-purpose corpus, is refined through a trainable intermediate layer. We further learn an image-conditioned latent representation from the tagging space and an image-conditioned latent representation from the ground-truth caption space. By aligning the image-conditioned tagging space with the image-conditioned caption space, we refine image and vocabulary tagging representations, thereby improving region-level caption generation performance. Additionally, different from traditional approaches that directly concatenate image features and keyword embeddings as a latent query and feed them into a frozen LLM decoder, we address the alignment challenge between multimodal inputs and frozen LLMs. Since LLMs are pre-trained in unimodal settings and have not been exposed to image data, freezing LLMs during multimodal training complicates vision-language alignment. To resolve this, we introduce a  \textbf{\textit{semantic space alignment}} module that uses contrastive learning. This module aligns the combined representation of the image’s latent features and the tag embeddings with the embeddings of the ground truth captions. The tags are embedded using the LLM’s tokenizer and embedding layer. Similarly, the ground truth captions are also encoded by the LLM’s tokenizer and embedding layers. This ensures compatibility between the multimodal features and the frozen LLM, thereby improving the quality of multimodal representations and overall caption generation performance.

Let $\mathcal{L}_{\text{AlignCap}}$ represent the overall loss of the model. Here, $\mathcal{L}_{\text{tag}}$ refers to the tagging loss applied to the region tagging module, $\mathcal{L}_{\text{cap}}$ denotes the captioning loss incorporated into the LLM module, $\mathcal{L}_{\text{cond}}$ corresponds to the alignment loss aligning the image-conditioned latent representations, and $\mathcal{L}_{\text{multi}}$ represents the loss  used to refine the multi-modal representations before feeding them into the LLM. The hyperparameters $\alpha$, $\beta$, $\gamma$, and $\lambda$ act as weighting coefficients balancing the contributions of each loss term.

The overall loss function is defined as:
\begin{equation}
\mathcal{L}_{\text{AlignCap}} = \alpha \cdot \mathcal{L}_{\text{tag}} + \beta \cdot \mathcal{L}_{\text{cap}} + \gamma \cdot \mathcal{L}_{\text{cond}} + \lambda \cdot \mathcal{L}_{\text{multi}}.
\end{equation}


\subsection{Visual Feature Extraction Module}

\begin{figure*}[htbp!]
  \centering
   \includegraphics[width=1\linewidth]{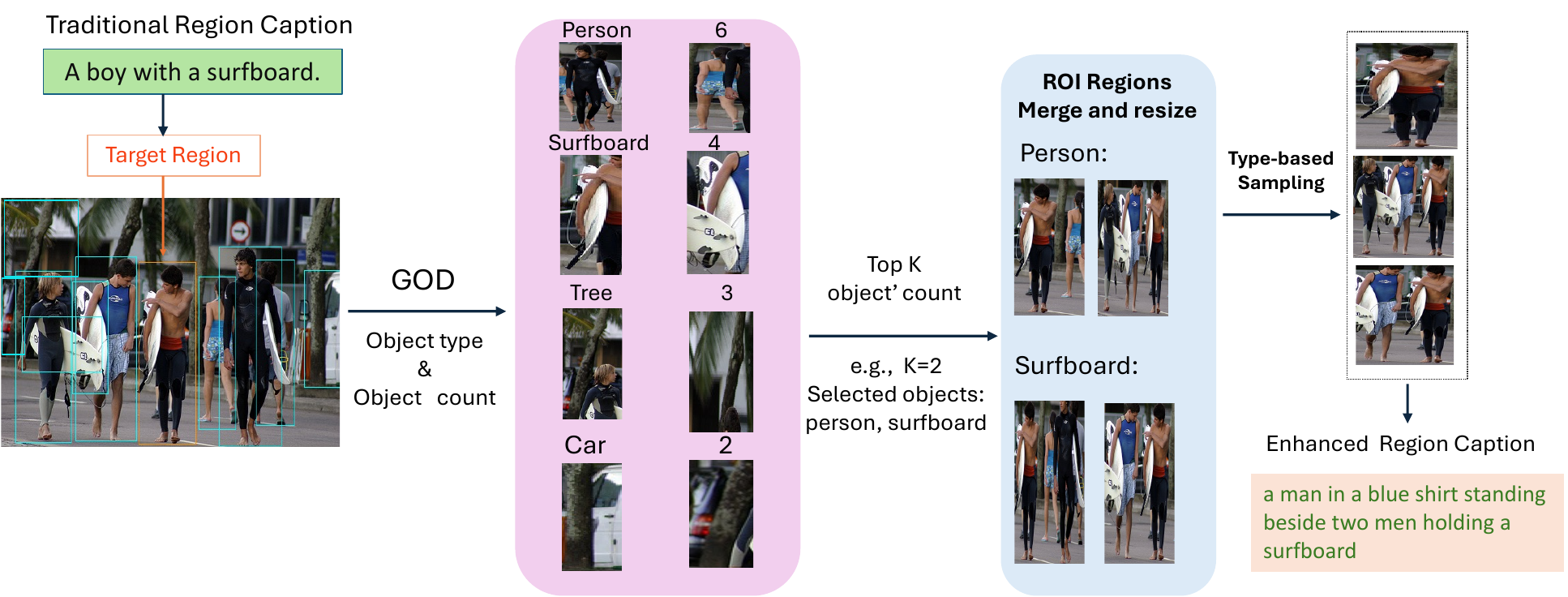}
\caption{Illustration of the GOD module. First, it proposes general object localizations. Then, it combines the selected target region with the proposed objects and performs cropping. Finally, it samples the required number of views needed for region captioning.}
   \label{fig:god_mod}
   \vspace{-3.5mm}
\end{figure*}

We begin by utilizing the GOD module (Fig.~\ref{fig:god_mod}) to detect objects of interest, classifying them into their corresponding base classes. The detected objects are then ranked by object frequency for each class, from which we select the top \( k \) classes. For example, given detected objects \(\{\text{person}: 4, \text{dog}: 3, \text{mouse}: 2\}\), with \( k = 1 \), "person" would be the top class. After determining the top \( k \) classes, the areas of each object from the selected class are merged with the target region, where the boundary is determined by either the minimum or maximum area values, forming a new candidate view. For model optimization, \( j \) candidate views are randomly sampled from each class during training. At inference time, we select the candidate view with the greatest discrepancy from the target region, calculated using the method from  to assess visual differences between images.

The input that contains 1 target region and j-1 candidate views will be sent to the frozen VIT encoder to first sample the corresponding features of each view. Then, a spatial awareness module will be used to align these features from different views. We first use a RoI-align module  to extract features better aligned to the proposed region of interest. We employ a trainable cross-attention module to align features from distinct spatial contexts. The module takes the candidate view embeddings and target region embeddings as input. Initially, both embeddings are processed through a LayerNorm layer. Subsequently, the embeddings are fed into a multi-head cross-attention layer, where the key, query, and value are defined as follows: the candidate view embedding serves as the query, retrieving the most relevant regions from the target view embedding, which acts as the key. The value is also derived from the candidate view embedding, ensuring that the attention mechanism refines region-level features based on the interaction between the candidate and target views. Following the cross-attention, dropout is applied, and the result is combined with the original candidate view embeddings via a residual connection. The output undergoes an additional LayerNorm step and is then passed through an MLP with two hidden layers and a SiLU activation function to extract the visual features based on the target view and candidate views' cross-attention. Finally, the MLP output is added to the previous result, producing the final output.

We apply a trainable \textbf{alignment module} that processes visual features learned from the cross-attention module, generating latent queries that directly engage with the LLM to enable text generation conditioned on visual input. Additionally, we refine these latent representations to achieve better alignment between different modalities, as will be discussed in the following section.

\subsection{Latent Feature Refinement Module } 


We propose a Latent feature refinement module (detailed in Appendix Fig.) to refine the latent representation by using a frozen CLIP encoder to encode the vocabulary into textual label queries. A linear layer acts as an adapter to refine the coarse-grained vocabulary latent features from the frozen encoder, mapping them into a space compatible with the multi-modal framework. To learn a fine-grained representation, we first send the latent features of the image and text to a \textbf{latent feature extraction module}, where we apply linear projection layers to both latent image and tagging space features, aligning them for cross-attention. The image features query the coarse-grained vocabulary tagging features to generate image-conditioned tagging space features. Additionally, we extract ground-truth caption features via a frozen CLIP encoder, applying projection layers to both image and ground-truth caption features to produce image-conditioned caption space features. We then send these features to a \textbf{fine-grained contrastive learning module}. A two-layer cross-attention mechanism is employed to learn a similarity score matrix, which is optimized using an objective function inspired by .

Formally, given a batch of \( N \) image-conditioned tagging space features \( \mathcal{T} = [t_1, t_2, \dots, t_N] \) and their corresponding image-conditioned caption space features \( \mathcal{C} = [c_1, c_2, \dots, c_N] \), we compute pairwise similarity score matrix. This results in \( N \times N \) similarity matrices, where each element \( s_{ij} \) is learned via the output of the module. The module aligns conditioned tagging and caption features by generating attention-weighted representations, followed by normalization and projection layers, producing a similarity matrix for the final scores. According to , the loss function is defined as:

\begin{equation}
L_{cond} = - \log \left( \frac{1}{1 + \exp(Z_{ij} \cdot (- \tau \cdot s_{ij} + b))} \right),
\label{eq:conditional_loss}
\end{equation}
where \( Z_{ij} \) is the label indicating whether the pair \( (t_i, c_j) \) is matching (\( Z_{ij} = 1 \)) or non-matching (\( Z_{ij} = -1 \)). \( s_{ij} \) is the similarity score learned from the module between the embeddings \( c_i \) and \( t_j \). \( \tau \) is a learnable temperature parameter that scales the logits, and \( b \) is a bias term that compensates for the imbalance between positive and negative pairs. This loss encourages high similarity for matching pairs while penalizing non-matching pairs, thus aligning the embedding spaces effectively.

\subsection{Semantic Space Alignment}

As contrastive learning on latent embeddings has demonstrated considerable potential for enhancing downstream tasks , we aim to refine the final embeddings before passing them to the LLM. Initially, we use the LLM's tokenizer and encoder to embed the recognized tags. Typically, the LLM processes the concatenated tag and image embeddings. To ensure alignment with the ground truth semantic space, the semantic space alignment module (detailed in Appendix Fig.) aligns the input embeddings with the ground truth caption embeddings by embedding the captions using the LLM's tokenizer and encoder. The concatenated image and tag embeddings are then sent to the \textbf{LLM embedding alignment module}, where the latent features are projected through a linear projection layer. This process helps learn a unified multi-modal space encapsulating both visual and linguistic features, providing necessary contextual information. This unified embedding is further processed by a \textbf{fine-grained contrastive learning module}, which uses a two-layer cross-attention mechanism with the embedded captions. The output is normalized and passed through a projection layer, resulting in a similarity matrix for contrastive learning. Finally, we optimize the learning process using a contrastive loss function to ensure that the unified input embeddings are semantically meaningful and suitable for use in the LLM.

Mathematically, given a batch of \( N \) image features and their corresponding text features, we compute their joint embedding features \( \mathcal{E} = \{\mathbf{e}_1, \mathbf{e}_2, \dots, \mathbf{e}_N\} \) and ground-truth caption embedding \( \mathcal{C} = \{\mathbf{c}_1, \mathbf{c}_2, \dots, \mathbf{c}_N\} \). We calculate pairwise similarity score matrices using the same method as in Equation ~\ref{eq:conditional_loss}. Accordingly, the loss function is defined as:

\vspace{-3.5mm}

\begin{equation}
L_{multi} = - \log \left( \frac{1}{1 + \exp(Z_{ij} \cdot (- \tau \cdot s_{ij} + b))} \right).
\end{equation}

Unlike Equation~\ref{eq:conditional_loss}, in this context, \( Z_{ij} \) denotes the label indicating whether the pair \( (\mathbf{e}_i, \mathbf{c}_j) \) is a match (\( Z_{ij} = 1 \)) or a non-match (\( Z_{ij} = -1 \)).
\section{Experiment}

\end{document}